\documentclass{article}%
\usepackage{amsmath}%
\usepackage{amsfonts}%
\usepackage{amssymb}%
\usepackage{graphicx}
\usepackage{algorithm}
\usepackage{algorithmic}
\usepackage{appendix}
\pdfoutput=1
\begin{document}

\title{D-Bees: A Novel Method Inspired by Bee Colony Optimization for Solving Word Sense Disambiguation}

\author{
	Sallam Abualhaija, Karl-Heinz Zimmermann \\ 
}

\date{May 6, 2014}
\maketitle

\begin{abstract}
Word sense disambiguation (WSD) is a problem in the field of computational linguistics 
given as finding the intended sense of a word (or a set of words) when it is activated within a certain context. 
WSD was recently addressed as a combinatorial optimization problem in which the goal is to find a sequence of senses 
that maximize the semantic relatedness among the target words. 
In this article, a novel algorithm for solving the WSD problem called D-Bees is proposed which is inspired by bee colony optimization (BCO)
where artificial bee agents collaborate to solve the problem. 
The D-Bees algorithm is evaluated on a standard dataset (SemEval 2007 coarse-grained English all-words task corpus) 
and is compared to simulated annealing, genetic algorithms, and two ant colony optimization techniques (ACO). 
It will be observed that the BCO and ACO approaches are on par.
\end{abstract}

\newpage

\section{Introduction}
Word sense disambiguation (WSD) is a problem in the field of computational linguistics 
defined as finding the intended sense of a word (or a set of words) when it is activated within a certain context (Agirre and Edmonds 2006). 
For example, in the sentence \emph{"I bought a new wireless mouse for my Apple Mac laptop"}, 
\textit{mouse} means a computer device and not a rodent while \textit{apple} refers to the computer company sense and not to a fruit.

WSD is a difficult task for a machine to solve due to the fact that not all words are mono-sensed,
rather they may have several meanings varied with the context in which they occur. 
Words are called \textit{homonymous} if they have several distinct meanings, 
e.g.,\ \emph{bank} could mean the financial institution or the side of a river, 
and \textit{polysemous} if the meanings are related, 
e.g.,\ \emph{bank} could refer to the financial institution with its logical meaning or the physical building based on the context. 
The question to which level a word should be disambiguated, i.e.,\ how specific senses should be, is application dependent; 
since WSD is usually not a stand-alone problem, but integrated within other applications like machine translation (see Vickrey et al. 2005) or information retrieval (see Sanderson 1994) each of which require different levels of distinction. 

The straightforward method to tackle WSD problem is to find all the senses of each word in the text 
and compare them with the senses of all other words within a certain context window. 
Thus reporting the sense which provides a maximum overall relatedness to the other potential senses. 
However, this straight forward method is not practical because the time complexity increases exponentially with the size of the context. 
The problem is NP-complete (Agirre and Edmonds 2006), the larger the size of the context window the sooner we get a combinatorial explosion, and the time needed to solve it increases exponentially. 

WSD is still an open research problem although it is as old as machine translation due to the widely available massive amount of texts that are increasing drastically by time. 
Hence, finding efficient text processing tools and systems to facilitate communication, for which WSD is considered as a backbone step, becomes a task beneath a spot light. 
Initially, WSD was considered as a classification task (Agirre and Edmonds 2006) where word senses are the classes and the system should assign each occurrence of a word to one or more appropriate 
senses (classes).
Correspondingly, supervised approaches were introduced to solve the problem by using machine learning methods, such as naive Bayesian (see Pedersen 2000), 
to induce a classifier based on available annotated corpora. 

An annotated corpus is usually created by defining correct meanings of each occurrence of a word manually. 
After this, these annotated corpora become the examples used to train classifiers which are then used to classify new occurrences of the same words as in the samples. 
It is clear that the more training samples are available, the better the performance of the classifier. 
Moreover, the senses of words could be retrieved automatically from a machine readable dictionary (MRD) such as the well-known WordNet.
WordNet is a lexical database that contains $155.000$ words organized in more than $117,000$ synsets (Miller 1995). 
A synset is the main component in WordNet representing synonyms that form together a certain meaning. 
The meaning of a synset is given as a definition.  
The process of creating annotated corpora is not only exhausting but also necessary for each language. 
Moreover, active languages evolve by time such that even more effort is needed to get new examples if new terms appeared suddenly or vanished.
For instance, the word \emph{``rock''} nowadays has the meaning of a stone as well as music genre. 
To avoid being entrapped in the problem of preparing annotated corpora, attention needs to be paid to new approaches and perspectives in the knowledge-based unsupervised direction, 
one of the recent trends to address WSD as a combinatorial optimization problem. 

In any optimization problem, a cost function called the objective function is to be optimized given a set of feasible solutions, which are the solutions or elements of a universe that
satisfy the constraints.
From the WSD perspective, the objective function is the relatedness measure between two senses and the goal is to attain the senses which maximize the overall relatedness value. 
One of the well known measures, which is intuitive and uses the definitions of the senses from a dictionary, is the Lesk algorithm 
in which the similarity value is calculated by counting the overlapping words between two definitions of the senses (Lesk 1986). 
The Lesk algorithm has been extended by Banerjee and Pedersen (2002) such that instead of considering only the immediate definitions of the senses in question, 
the semantically related senses are also taken into account, like hypernyms, hyponyms and others, leading to a more accurate similarity value. 
In order get the senses' definitions, any sense inventory could be used such as WordNet. 

WSD can be defined as an optimization problem (Pedersen, Banerjee and Patwardhan 2005).
For this, let $C=\{w_1, w_2, ..., w_{n}\}$ be a set of $n$ words given by a window of context of length $n$.
Let $w_{t}$ be the target word to be disambiguated, $1\leq t\leq n$.
Suppose each word $w_{i}$ has $m$ possible senses $s_{i1}, s_{i2}, ..., s_{im}$, $1\leq i\leq n$.
Then the objective function is
\begin{equation} \label{Opt}
{\rm argmax}_{i=1}^m \sum_{j=1}^{n}{\max \{ {\rm rel}(s_{ti}, s_{j1})},\ldots,{\rm rel}(s_{ti}, s_{jm})\}, 
\end{equation}
where ${\rm rel}$ is the relatedness value between two senses.
The task is then to find a sequence of senses which maximizes the overall relatedness value among the words within a certain context window of length $n$.
The overall relatedness is calculated for each sequence and finally the sequence that resulted in the best relatedness is considered. 

In addition to the brute force method (Pedersen, Banerjee and Patwardhan 2005) initially proposed to tackle this problem, 
several bio-inspired techniques have been proposed to optimize the cost function, like simulated annealing (see Cowie, Guthrie and Guthrie 1992), genetic algorithms (Zhang, Zhou and Martin 2008), 
and ant colony optimization (see Schwab and Guillaume 2011), (see also Nguyen and Ock 2011). 

This article introduces D-Bees, a novel knowledge-based unsupervised method for solving WSD problem which has been inspired by bee colony optimization (BCO). 
In the following, the BCO meta-heuristic is first discussed in general.
Then the D-Bees method is described and  after that experiments and results are illustrated and compared to the previous methods. 
Moreover, a pseudo code of the D-Bees algorithm can be found in the appendix.

\section{Bee Colony Optimization}
There are several proposed computational methods inspired by honey bees in nature 
each of which used in a certain application. 
In this paper, we have adapted the bee colony optimization (BCO) meta-heuristic which was first proposed by Teodorovi{\'c} (2009). 

Social insects in general are self-organized and adapt well to the environmental changes.
This is usually facilitated by exchanging information among the individual insects in order to achieve a collective intelligence (emergence) for the sake of the colony. 
Unlike ants that interact indirectly by depositing a chemical substance along the path called \textit{pheromone}, 
bees interact directly by performing a sort of dance on a dancing floor in the hive. 

First, bee scouts explore the unknown environment looking for a food resource from which they can collect nectar for the hive. 
Once a food source has been found, they head back to the hive and perform a certain dance based on the goodness of the food resource and the distance to it
which amounts to an advertisement or recruit to other bee fellows to further exploit this food resource. 
There are two types of dances, a round dance if the food source is close to the hive, and a waggle dance if the food is farther away, through which the bees also give information about the direction to the food source.

Having watched the dance floor, the uncommitted bees may decide to follow one of the advertised paths. 
The committed bees can stick to their own path or abandon it and follow one of the other advertised paths. 
These decisions usually depend on the hive needs and the characteristics of the food resources like its goodness.

The computational BCO assumes that each bee agent explores part of the search space of the combinatorial problem and generates a particular solution of the problem. 
For this, the number of bee agents are predefined.
The process is simulated by two alternating phases, a forward pass and a backward pass. 
In a forward pass, a bee agent travels a number of steps which is predefined based on the problem. 
In a backward pass, all bee agents return back to the hive and exchange information among them indicating the goodness of the sub-solution and the partial path found. 
Each bee agent decides with a certain probability as described in Eq.~(\ref{loyal}) whether to stay loyal to its own path or to abandon it. 
The bee agents with the best found solutions are more likely to be loyal to their paths and therefore become recruiters advertising their partial solutions. 
However, there is always a slight chance for a bee agent to stick to its own path even though it might be not good enough hoping that this path might finally lead to a better solution.

This chance will get smaller by time, i.e.\ the larger the number of forward passes, the less the chance for bee agents to abandon their paths.

The {\em loyalty probability\/} of the $b$-th bee agent is given by the negative exponential function (Teodorovi{\'c} 2009)
\begin{equation}
p_b^{u+1} = e^{-\frac{O_{\max} - O_b}{u}}
\label{loyal}
\end{equation}
where $u$ is the number of the forward passes made so far, $0\leq u\leq n$, 
$O_b$ is the normalized value for the objective function of the partial solution created by the $b$-th bee, 
and $O_{\max}$ is the maximum overall normalized value of the partial solutions.

Furthermore, the bee agents that have abandoned their paths select one of the advertised solutions.
This is given by the {\em recruiting probability\/} of the $b$-the bee agent (Teodorovi{\'c} 2009) 
\begin{equation}\label{recruit}
p_b = \frac{O_b}{\sum_{k=1}^{R}{O_k}}
\end{equation}
where $R$ indicates the number of recruiters and $O_k$ represents the normalized value for the objective function of the $k$-th  advertised partial solution.

The forward and backward passes are alternated until bee agents generate feasible solutions. 
This process is repeated until the maximum number of iterations is reached or the solution cannot be improved any further.  
A pseudo code for the BCO meta-heuristic is given by Teodorovi{\'c} (2009).

\section{D-Bees}
D-Bees is a knowledge-based unsupervised method adapting the BCO meta-heuristic to solve the WSD problem. 
Given a set of target words as input, 
the system finds a corresponding sequence of senses that are likely intended by the target words. 
In a pre-processing stage, the target words are ordered based on their part of speech (POS). 
The Lin measure is used to calculate the similarity between two senses if they have similar POS, while a normalized version the Lesk measure is used otherwise.

The Lin measure (Lin 1997) is based on the \textit{information content\/} (IC) of a concept which measures how specific a particular concept in a certain topic is.
The value of IC is calculated by counting the frequency of the concept in a large corpus determining the probability of its occurrence by maximum likelihood estimation. 
The Lin measure calculates the relatedness between two concepts as the ratio of the IC of their \textit{lowest common subsumer\/} (LCS).

At first, a random target word is chosen to represent the hive whereas the other target words represent the food resources from which the bee agents collect information. 
The number of bee agents is given by the number of senses of the target word and each bee agent holds one of the sense definitions.
Moreover,  the quality of each path that is initially set to zero.

In a forward pass, each bee evaluates the next move by calculating the similarity value between the sense that the bee currently holds 
and a random sense chosen from the set of senses of the next word. 
Yet, the bee agents choose the sense which leads to the maximum similarity value. 
After updating the current sense and the quality by incrementally adding the similarity values together, the bee agent moves a step further 
until the number of constructive moves (NC) is reached. 

After partial solutions have been found, the bee agents return to the hive, exchange information with each other and initiate the backward pass. 
For this, each bee agent calculates the loyalty probability as in Eq.~(\ref{loyal}) and then decides whether to stay loyal to its path or to become uncommitted and follow one of the advertised solutions. 
The bee agents holding the best three solutions in terms of quality advertisement are then followed by the uncommitted bee agents using Eq.~(\ref{recruit}).

The forward and backward passes are alternated until there are no more target words to disambiguate. 
The bee agent with the best solution found in terms of quality is stored as a potential solution. 
The algorithm is iterated until the maximum number of iterations is reached or there is no significant improvement on the previously found solution.
In our experiments, ten iterations will be made and the quality of each path is evaluated by a threshold $\beta$ that is set to 0.8. 
Finally, the best solution is returned as an output.

\begin{figure}[h!]
	\includegraphics[scale=0.77,bb = 0 0 300 300]{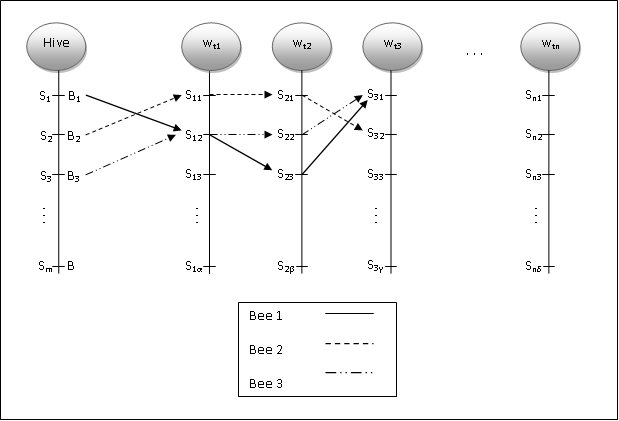}
	\caption{An illustration of the forward pass.}
	\label{img}
\end{figure}

Figure~\ref{img} illustrates the principles of the forward and backward pass. 
The hive represents a random target word and the nodes $1,\ldots,n$ are the food resources which represent the rest of the target words. 
The bee agents move among the target words by choosing an appropriate sense as explained above. 
Note that each word may have a different number of senses.
The algorithm is designed to disambiguate a set of target words.
It could also be customized to solve lexical substitution.

\section{Experiments and Results}
The system is tested on the SemEval 2007 coarse-grained English all-words task corpus (Navigli, Litkowski and Hargraves 2007). 
The task is composed of five different texts where the first three are obtained from the Wall Street Journal corpus, the fourth is a Wikipedia 
article about computer programming,
and the last is an excerpt of Amy Seedman's \textit{Knights of the Art\/} biography (Navigli, Litkowski and Hargraves 2007).

Table \ref{semEvalData} illustrates the domains addressed by these texts and the distribution of words as described in the texts (Navigli, Litkowski and Hargraves 2007). 
\begin{table}[h!]
\centering
\begin{tabular}{|c|c|c|c|}
		\hline
		Article & Domain & Words & Annotated\\
		\hline
		d001 & Journalism & 951 & 368\\
		\hline
		d002 & Book Review & 987 & 379\\
		\hline
		d003 & Travel & 1311 & 500\\
		\hline
		d004 & Computer Science & 1326 & 677\\
		\hline
		d005 & Biography & 802 & 345\\
		\hline
\end{tabular}
	\caption{The five articles in the dataset.}
	\label{semEvalData}
\end{table}

Python 2.7 has been used to implement the system along with NLTK (Bird, Klein and Loper 1992).
The experiments were conducted on an Intel PC i5-2450M CPU 2.50GHz. 
WordNet has been integrated to NLTK to get the senses of the target words and their definitions along with the benefit of the semantic relations, 
such as hyponymy, hypernymy, and so on. 

Furthermore, the evaluation criteria are \emph{attempted\/} which indicates how many words the system can disambiguate, 
\emph{precision} which measures how many target words are correctly disambiguated and so gives the accuracy of the system, 
\emph{recall} which is defined by the ratio between the number of correctly disambiguated target words and the total number of the target words in the dataset,
and the \emph{F-measure} which is the harmonic mean of the precision and recall values as described in the following equation
\begin{equation}
\mbox{F-measure} = 2\cdot \frac{\mbox{precision}\cdot\mbox{recall}}{\mbox{precision}+\mbox{recall}}.
\label{f}
\end{equation}

%
%

The D-Bees algorithm is parametrized by the number of bees that are produced in a hive which corresponds to the number of the senses,
the number of constructive movements in a forward pass which is set to 3, 
the number of recruiters $R$ that is also set to 3, 
the maximum number of iterations is set to 10, 
and the quality of each path evaluated by a threshold $\beta$ which is set to 0.8. 

Based on these parameters, the D-Bees algorithm has achieved the results given in Table~\ref{detailed_Results}. These results represent a single run; due to the high time complexity. 
\begin{table}[h!]
\centering
\begin{tabular}{|c|c|c|c|}
		\hline
		Text  & Precision(\%) & Recall(\%) & F-measure(\%)\\
		\hline
		d001 & 83.42 & 83.42 & 83.42\\
		\hline
		d002 & 84,52 & 84,52 & 84,52\\
		\hline
		d003 & 80,16 & 80,16 & 80,16\\
		\hline
		d004 & 78,33 & 78,15 & 78,24\\
		\hline
		d005 & 75,95 & 75,78 & 75,86\\
		\hline
\end{tabular}
	\caption{D-Bees performance on each article in the dataset.}
	\label{detailed_Results}
\end{table}

Obviously, the precision of the D-Bees algorithm is better for the first three texts and thus behaves similar to other systems applied on the same dataset (Navigli, Litkowski and Hargraves 2007). 
The last two texts are more domain specific which might explain the reason for attaining lower precision values. 
It follows that the current D-Bees algorithm is more suitable for disambiguating general texts.

The results of the D-Bees algorithm have been compared with other optimization methods, like simulated annealing (SA), genetic algorithms (GA), 
and two ant colony optimization techniques ACA (Schwab et al. 2011) and TSP-ACO (Nguyen and Ock 2011). 
The upper-bound is the inter-annotator agreement which is approximately 86.44\% (Navigli, Litkowski and Hargraves 2007). 
Moreover, two baselines were provided,  namely, a most frequent sense (MFS) system that has achieved 78.89\% 
and a random sense (RS) system that has attained 52.43\%. 
Table~\ref{results} summarizes the results.
\begin{table}[h!]
\centering
\begin{tabular}{|c|c|c|c|c|}
		\hline
		System & Attempted(\%) & Precision (\%) & Recall (\%) & F-Measure (\%) \\
		\hline
		\textbf{D-Bees} & 99.91 & \textbf{80.47} & \textbf{80.41} & \textbf{80.44} \\
		\hline
		MFS & 100 & 78.89 & 78.89 & 78.89 \\
		\hline
		TSP-ACO & 99.80 & 78.50 & 78.10 & 78.30 \\
		\hline
		ACA & 100 & 77.64 & 77.64 & 77.64 \\
		\hline
		SA & 100 & 74.23 & 74.23 & 74.23 \\
		\hline
		GA & 100 & 73.98 & 73.98 & 73.98 \\
		\hline
		RS & 100 & 52.43 & 52.43 & 52.43\\
		\hline
\end{tabular}
	\caption{Comparison of D-Bees with other methods.}
	\label{results}
\end{table}

In our study, the D-Bees algorithm has achieved competitive results to the other algorithms. 
In particular, the genetic algorithm and simulated annealing have attained the worst results since they are computationally very intensive and non-adaptive.
Here swarm intelligence techniques have led to better results since the agents can maintain their memories about partial solutions.
Moreover, they can communicate with each other and exchange knowledge regarding the goodness of partial solutions. 
Therefore, these algorithms find solutions in a more efficient way.


Bee colony optimization is up on par with both ant colony optimization techniques.
Unlike ACO, in which ant agents follow the pheromone values on a trail and choose the path with the highest amount of pheromone, 
bee agents evaluate different sub-paths every time they get back to the hive according to the quality of these paths. 
This enables them to emphasize on promising solutions and neglect the worse solutions efficiently. 
Moreover, the direct communication among bee agents, through the waggle dances, gives a better possibility for the uncommitted bees to choose from 
and follow with a certain probability the promising sub-paths based mainly on their quality. 
Both ACO and BCO have the advantage to easily adapt to a dynamic environment which is important for the WSD problem since the natural languages quickly evolve.

\section{Conclusion}
In this paper, the D-Bees algorithm has been introduced, a novel knowledge-based unsupervised method for solving the problem of WSD inspired by bee colony optimization. 
The experiments on the standard dataset SemEval 2007 coarse-grained English all-words task corpus have shown that D-Bees achieves promising results 
and competitive to the other methods in this field. 
This encourages further research work on D-Bees and related algorithms.



\section{References}

\begin{itemize}
\setlength{\itemindent}{-.3in}

\item[] Agirre, Eneko, and Edmonds, Philip. 2006. \textit{Word Sense Disambiguation: Algorithms and Applications}. Springer. 
\item[] Banerjee, Satanjeev, and Pedersen, Ted. 2002. An Adapted Lesk Algorithm for Word Sense Disambiguation using WordNet. In \textit{Computational linguistics and intelligent text processing}, pages 136--145. Springer Berlin Heidelberg. 
\item[] Bird, Steven, and Klein, Ewan and Loper, Edward. 2009. \textit{Natural Language Processing with Python}. O'Reilly Media, Inc.. 
\item[] Cowie, Jim, Guthrie, Joe, and Guthrie, Louise. 1992. Lexical Disambiguation Using Simulated Annealing. In \textit{Proceedings of the 14th conference on Computational linguistics-Volume 1}, pages 359--365. Association for Computational Linguistics.
\item[] Lesk, Michael. 1986. Automatic Sense Disambiguation using Machine Readable Dictionaries: How to Tell a Pine Cone from a Ice Cream Cone. In \textit{Proceedings of the 5th annual international conference on Systems documentation}, pages 24--26. ACM. 
\item[] Lin, Dekang. 1997. Using Syntactic Dependency as Local Context to Resolve Word Sense Ambiguity. In \textit{Proceedings of the 35th Annual Meeting of the Association for Computational Linguistics and Eighth Conference of the European Chapter of the Association for Computational Linguistics}, pages 64--71. Association for Computational Linguistics.
\item[] Miller, George A. 1995. WordNet: a Lexical Database for English. \textit{Communications of the ACM}, 38(11):39--41. 
\item[] Navigli, Roberto, and Litkowski, Kenneth C. and Hargraves, Orin. 2007. SemEval-2007 Task 07: Coarse-grained English All-words Task. In \textit{Proceedings of the 4th International Workshop on Semantic Evaluations}, pages 30--35. Association for Computational Linguistics. 
\item[] Nguyen, Kiem-Hieu, and Ock, Cheol-young. 2011. Word Sense Disambiguation as a Traveling Salesman Problem. \textit{Artificial Intelligence Review}, 40(4):pages 1--23. 
\item[] Pedersen, Ted. 2000. A Simple Approach to Building Ensembles of Naive Bayesian Classifiers for Word Sense Disambiguation. In \textit{Proceedings of the 1st North American Chapter of the Association for Computational Linguistics Conference}, pages 63--69. Association for Computational Linguistics.
\item[] Sanderson, Mark. 1994. Word Sense Disambiguation and Information Retrieval. In \textit{Proceedings of the 17th Annual International ACM SIGIR Conference on Research and Development in Information Retrieval}, pages 142--151. Springer-Verlag New York, Inc.. 
\item[] Schwab, Didier, and Goulian, J{\'e}r{\^o}me and Tchechmedjiev, Andon and Blanchon, Herv{\'e}. 2012. Ant Colony Algorithm for the Unsupervised Word Sense Disambiguation of Texts: Comparison and Evaluation. In \textit{COLING}, pages 2389--2404. 
\item[] Schwab, Didier, and Guillaume, Nathan. 2011. A Global Ant Colony Algorithm for Word Sense Disambiguation Based on Semantic Relatedness. \textit{Highlights in Practical Applications of Agents and Multiagent systems}, pages 257--264. Springer Berlin Heidelberg. 
\item[] Teodorovi{\'c}, Du{\v{s}}an. 2009. Bee Colony Optimization (BCO). In \textit{Innovations in swarm intelligence}, pages 39--60. Springer Berlin Heidelberg.
\item[] Vickrey, David, and Biewald, Luke and Teyssier, Marc and Koller, Daphne. 2005. Word-Sense Disambiguation for Machine Translation. In \textit{Proceedings of the Conference on Human Language Technology and Empirical Methods in Natural Language Processing}, pages 771--778. Association for Computational Linguistics. 
\item[] Zhang, Chunhui, Zhou, Yiming, and Martin, Trevor. 2008. Genetic Word Sense Disambiguation Algorithm. \textit{Intelligent Information Technology Applications}, Volume 1, pages 123--127. IEEE.
\end{itemize}

\newpage
\appendix 

\section{D-Bees Algorithm}
The parametrization of the D-Bees algorithm is as follows:

\begin{itemize}
	\item $B$: The number of bee agents in the hive
	\item hive: A target word that is chosen randomly 
	\item NC: Number of constructive movements
	\item $\beta$ : Quality threshold that controls the maximum number of iterations
	\item $\theta$ : Similarity threshold that controls the similarity value
	\item $R$: The number of recruiting bees
\end{itemize}

\newpage
\begin{algorithm}[H]
\caption{D-Bees pseudo code}

	\begin{algorithmic}[1]
		\STATE $\beta \leftarrow 0.8$
		\STATE $\theta \leftarrow 0.5$
		\STATE hive $\leftarrow w_{t}$ \COMMENT{$w_{t}$ is a randomly chosen target word}
		\STATE $B$ $\leftarrow$ number of senses in the hive 
		\STATE NC $\leftarrow 3$
		\STATE $R$  $\leftarrow 3$  
		\STATE maxIterations $\leftarrow 10$
		\STATE $u$ $\leftarrow$ 1 \COMMENT{$u$ is the number of the forward pass}
		\STATE bestSolution $\leftarrow$ Null
 		\REPEAT
			\STATE initializeBees() \COMMENT{each bee starts her path from a sense of the hive, with initial quality=0.0}
			\REPEAT
				\FORALL {$b_{i} \in$ $B$}
					\STATE nextSense $\leftarrow$ evaluate constructive moves \COMMENT{by calculating the similarity with random senses of the next $w_{t}$, until 5 senses, or until similarity $\geq \theta$}
					\STATE updateBee() \COMMENT{updating path and quality of the bee accordingly and move one step forward}
				\ENDFOR
				\STATE sortBees()
				\FORALL {$b_{k} \in$ $R$} 
					\STATE pRecruitment $\leftarrow \frac{O_{k}}{\sum^{R}_{j=0}{O_{j}}}$
				\ENDFOR
				\STATE weightedRecruiters $\leftarrow$ weighted list of recruiters based on pRecruitments \COMMENT{recruiters are bees with best partial solutions in terms of quality}
				\FORALL {$b_{i} \in$ $B$}
					\STATE pLoyalty $\leftarrow e^{-\frac{O_{max}-O_{i}}{u}}$ \COMMENT {probability of a $b_{i}$ being loyal to her path}
					\IF{\NOT loyal} 
						\STATE beeToFollow $\leftarrow$ random.choice(weightedRecruiters)
						\STATE $b_{i}$.path $\leftarrow$ beeToFollow.path  
					\ENDIF
				\ENDFOR
				\STATE $u \leftarrow u+1$
				\IF{bestSolution.quality $< b_{0}$.quality}
					\STATE bestSolution $\leftarrow b_{0}$
				\ENDIF
			\UNTIL{all words $w_{t}$ are visited}
		\UNTIL{path.quality $\leq \beta$ $\OR$ maxIterations}
		\RETURN bestSolution

	\end{algorithmic}
\end{algorithm}

\end{document}